\documentclass[conference]{IEEEtran}
\IEEEoverridecommandlockouts % This command is to use the \thanks command
\usepackage[T1]{fontenc}
\usepackage[cmex10]{amsmath}
\usepackage{amssymb}
\usepackage{graphicx}
\usepackage{balance}
\usepackage{multirow}
\usepackage{hhline}
\usepackage{makecell}
\usepackage{url}
\usepackage{cite}

\newcommand\set[1]{\mathcal{#1}}

\begin{document}
%% Paper meta-information
\title{\LARGE \bf
  A Proposal for Semantic Map Representation and Evaluation
}

\author{\IEEEauthorblockN{Roberto Capobianco, Jacopo Serafin, Johann
  Dichtl, Giorgio Grisetti, Luca Iocchi and  Daniele Nardi} % <-this % stops a space
  \IEEEauthorblockA{Department of Computer, Control and Management Engineering,\\
    Sapienza University of Rome, Italy
    \\{\tt\small \{lastname\}@dis.uniroma1.it}\\}%
}
%978-1-4673-9163-4/15/$31.00 ©2015 IEEE
% \IEEEpubid{\makebox[\columnwidth]{\hfill\bf
%     978-1-4673-9163-4/15/\$31.00~\copyright~2015
%     IEEE}\hspace{\columnsep}\makebox[\columnwidth]{}}

\maketitle

%%%%%%%%%%%%%%%%%%%%%%%%%%%%%%%%%%%%%%%%%%%%%%%%%%%%%%%%%%%%%%%%%%%%%%%%%%%%%%%%

\begin{abstract}
  Semantic mapping is the incremental process of ``mapping'' relevant
  information of the world (i.e., spatial information, temporal
  events, agents and actions) to a formal description supported by a
  reasoning engine. Current research focuses on learning the semantic
  of environments based on their spatial location, geometry and
  appearance. Many methods to tackle this problem have been proposed,
  but the lack of a uniform representation, as well as standard
  benchmarking suites, prevents their direct comparison. In this
  paper, we propose a standardization in the representation of
  semantic maps, by defining an easily extensible formalism to be used
  on top of metric maps of the environments. Based on this, we
  describe the procedure to build a dataset (based on real sensor
  data) for benchmarking semantic mapping techniques, also
  hypothesizing some possible evaluation metrics. Nevertheless, by
  providing a tool for the construction of a semantic map ground
  truth, we aim at the contribution of the scientific community in
  acquiring data for populating the dataset.
\end{abstract}

%%%%%%%%%%%%%%%%%%%%%%%%%%%%%%%%%%%%%%%%%%%%%%%%%%%%%%%%%%%%%%%%%%%%%%%%%%%%%%%%

\section{Introduction}

In the last years, semantic mapping has become a very active research
area. Such increasing interest is motivated by the idea that if robots
can \emph{understand} the environment in which humans live, and the
way they operate in it, they can also \emph{collaborate} and
\emph{act} (i.e., have a more cognitive behavior). Nevertheless, the
ability to \emph{communicate} represents a strict requirement for
collaboration among two or more agents. When dealing with humans, this
can be naturally achieved by enabling robots to use spoken language,
based on the learned semantic of the world. Associating symbols with
numerical representations in fact is a key requirement for producing a
robot that can use spoken language. Indeed, semantic mapping is the
incremental process of mapping relevant information of the world
(i.e., spatial information, temporal events, agents and actions) to a
formal description supported by a reasoning engine, with the aim of
learning to understand, collaborate and communicate.

\begin{figure}[t!]
  \centering
  \includegraphics[width=\columnwidth]{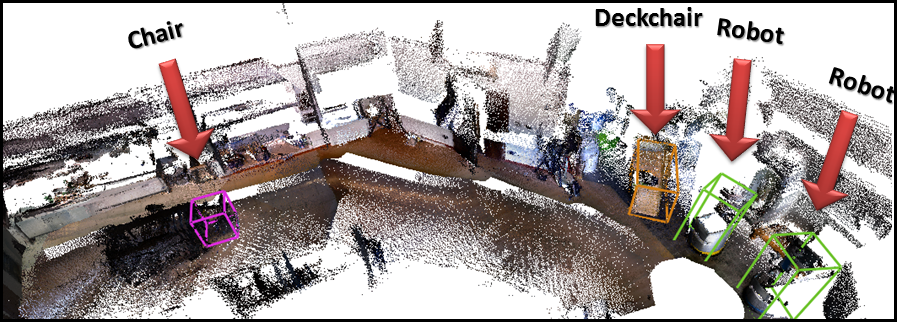}

  \vspace{0.1em}

  \includegraphics[width=\columnwidth]{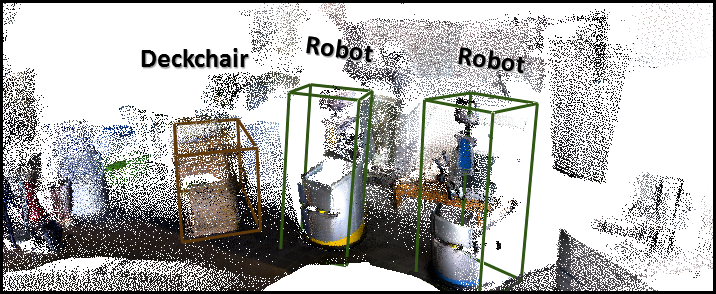}
  \caption{Double view of the example dataset acquired in the Robot
    Innovation Facility of Peccioli, in Italy. Part of the sitting
    room and the kitchen are shown, together with some bounding boxes
    identifying a chair, a deckchair and two robots.}
  \label{fig:teaser}
\end{figure}

Ongoing research mostly tries to address the problem by focusing on a
subset of the information to be learned, and by considering an agent
whose main abilities are navigation and object manipulation. In this
way, strict requirements for communicative or collaborative behaviors
are typically ignored. A relevant definition in this sense is given by
N{\"u}chter and Hertzberg \cite{Nuechter2008}, who describe a semantic
map for a mobile robot as ``\emph{a map that contains, in addition to
  spatial information about the environment, assignments of mapped
  features to entities of known classes. Further knowledge about these
  entities, independent of the map contents, is available for
  reasoning in some knowledge base with an associated reasoning
  engine}''. Based on the same concept, several approaches have been
proposed. These can be grouped in two main categories: fully automated
methods for classification of locations and objects~\cite{Blodow2011,
  Mozos2012, Gunther2013}, and techniques, which exploit the support
of the user in the knowledge acquisition and learning
process~\cite{Zender2008, Nieto-Granda2010, Pronobis2012}. While a
comprehensive overview of the relevant work in this direction can be
found in the survey by Kostavelis and
Gasteratos~\cite{Kostavelis2014}, it is important to remark that even
the simplest semantic map goes far beyond ``simple'' labeling of
spatial features. In fact, even though they are built on top of
sophisticate SLAM procedures, Computer Vision and Machine Learning
algorithms, semantic maps must provide the possibility to reason over
the acquired knowledge. Therefore they have to be formalized and
represented in a proper way. Moreover, semantic mapping methods cannot
be directly evaluated on the metrics and benchmarking datasets which
are available for other algorithms, since they do not take into
account any kind of reasoning. On the contrary, approaches proposed in
literature (Section~\ref{sec:related_work}) lack of any kind of
standardization and typically underestimate these questions. In
particular, two main issues emerge from the analysis of the
state-of-the-art: 1) the absence of a common formalism for
representing semantic maps and, consequently, 2) the lack of suitable
validation and evaluation techniques. This puts a significant
limitation on the research field, since it is difficult to understand
the improvements over the state-of-the-art and to even compare
available methods.

The aim of this paper is therefore twofold. First, we address the
above highlighted issues, by proposing a formalization and a
standardization in the representation of semantic maps (Section
\ref{sec:representation}). Second, we make a proposal for their
evaluation, as well as for benchmarking semantic mapping methods, by
means of a dataset based on real sensor data
(Section~\ref{sec:evaluation}). Moreover, by describing the procedure
and providing usable software\footnote[2]{The software is available at
  the following url: \url{http://goo.gl/v7xSyl}} for building such a
dataset (Section~\ref{sec:dataset}), we invite the scientific
community to contribute to its creation (see Fig.~\ref{fig:teaser} for
an example). Conclusions and open questions related to our proposal
are finally reported in Section~\ref{sec:discussion}.

%%%%%%%%%%%%%%%%%%%%%%%%%%%%%%%%%%%%%%%%%%%%%%%%%%%%%%%%%%%%%%%%%%%%%%%%%%%%%%%%

\section{Related Work}
\label{sec:related_work}

There exists a large literature on the problem of learning and
representing the semantics of environments based on their spatial
location, geometry and appearance~\cite{Kostavelis2014}. This activity
is usually referred to ``semantic mapping''. Such a term, although
originally describing a difficult process that deals with more
heterogeneous information (i.e., not limited to spatial knowledge),
has strong implications. Semantic maps should, in fact, not only
assign a certain number of labels or properties to relevant features
of the environment (like in~\cite{Goerke2009, Mozos2012}), but also
provide a representation of this knowledge in a form usable by the
system.

As introduced in the previous section, one of the main issues of
current research is the wide heterogeneity of the representations used
for semantic maps. For example Galindo \emph{et
  al.}~\cite{Galindo2005} represent environmental knowledge by
anchoring sensor data, that describe rooms or objects in a spatial
hierarchy, to the corresponding symbol of a conceptual hierarchy. Such
a conceptual hierarchy is based on a small ontology in description
logic, which enables the robot to perform inference. The authors
validate their approach by building their own domestic-like
environment and testing the learned model by executing navigation
commands. Pangercic \emph{et al.}~\cite{Pangercic2012}, instead,
investigate the representation of ``semantic object maps'' by means of
a symbolic knowledge base (in description logic) associated to Prolog
predicates (for inference). Such a knowledge base contains classes and
properties of objects, instances of semantic classes and spatial
information. While profiling the time required by the semantic mapping
process, the authors experiment their approach on a PR2 robot which
has to open a cabinet and to detect handles based on an apriori given
semantic map. Moreover, Bastianelli \emph{et
  al.}~\cite{Bastianelli2013} use a Prolog knowledge base containing
both the specific knowledge of a certain environment and the general
knowledge about a domain. The knowledge base is linked to the physical
environment by means of a matrix like data structure generated on top
of a metric map. Once again, the experimental validation is based on
qualitative evaluations of the robot behavior, given a certain command
and the learned semantic map. Riazuelo \emph{et
  al.}~\cite{Riazuelo2015} instead describe the RoboEarth cloud
semantic mapping system, which is composed of an ontology, for coding
concepts and relations, and a SLAM map for representing the scene
geometry and object locations. In particular, a recognition module
identifies objects based on a local database of CAD models, while the
whole system is integrated with an OWL ontology.

The other problem, which emerges as a consequence of the variety of
representations, is the absence of a standard suitable validation and
evaluation procedure. In addition to previous examples, Zender
\emph{et al.}~\cite{Zender2008} generate a representation ranging from
sensor-based maps to a conceptual abstraction, encoded in an OWL-DL
ontology of an indoor office environment. However, except for
individual modules, their experimental evaluation is mainly
qualitative. Pronobis and Jensfelt~\cite{Pronobis2012}, instead,
represent a conceptual map as a probabilistic chain graph model and
evaluate their method by comparing the robot belief to be in a certain
location against the ground truth. Gunther \emph{et al.}
\cite{Gunther2013} perform a sort of semantic aided object
classification based on an OWL-DL knowledge base. The evaluation is
based on the rate of correctly classified objects. Finally, Handa
\emph{et al.}~\cite{Handa2014} propose a synthetic dataset, which
could be eventually extended with semantic knowledge and used as a
ground truth for comparing semantic mapping methods. However, even
when noise is introduced, fictitious data never reflect a real world
acquisition.

Note that none of the cited works can compare the performance of their
semantic mapping method against those of other similar systems.
Starting from these considerations we propose a standard methodology
for representing and evaluating semantic maps. In particular, we
describe a formalization which includes a reference frame, spatial
information and a set of logic predicates. Such a formalization is
thought to be used as a general structure of the representation that
all the semantic maps have to include and can extend. Moreover, in
addition to proposing an evaluation metric, we suggest the procedure
for the creation of a semantic mapping dataset. In particular, such a
dataset is based on real sensor data enriched with semantic
information.

%%%%%%%%%%%%%%%%%%%%%%%%%%%%%%%%%%%%%%%%%%%%%%%%%%%%%%%%%%%%%%%%%%%%%%%%%%%%%%%%

\section{Semantic Map Representation}
\label{sec:representation}

As previously stated, in order to define a map to be ``semantic'', we
require that knowledge is represented in a suitable manner. In fact,
this enables additional information to be inferred from the map,
whenever a reasoning engine is associated to it. For this reason, in
this section, we propose a formalization of a \emph{minimal} general
structure of the representation that should be implemented in a
semantic map. This representation has to play the role of common
interface among all the semantic maps, and can be easily extended or
specialized as needed.

\begin{figure}[t!]
  \centering
  \includegraphics[width=\columnwidth]{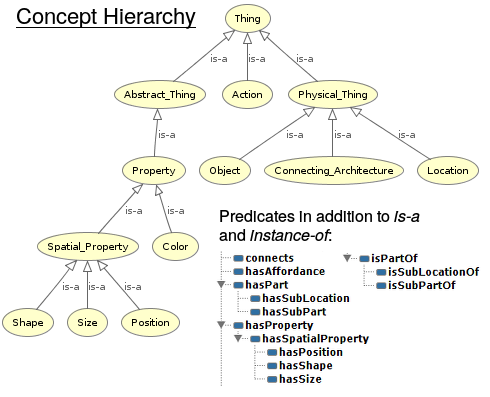}
  \caption{Minimal concept hierarchy to be used for a standard
    semantic map representation.}
  \label{fig:ontology}
\end{figure}

In the general formalization that we are describing, such a
representation is defined as a triple
\begin{equation}
  \set{SM} = \langle R,\set{M},\set{P} \rangle,
\end{equation}
where:

\begin{itemize}
\item $R$ is the global reference system in which all the elements of
  the semantic map are expressed;
\item $\set{M}$ is a set of geometrical elements obtained as raw
  sensor data. They are expressed in the reference frame $R$ and
  describe spatial information in a mathematical form. $\set{M}_s
  \subseteq \set{M}$ is the subset of semantically relevant elements;
\item $\set{P}$ is a set of predicates, among which
  \textit{is-a}(\texttt{X}, \texttt{Y}) and
  \textit{instance-of}(\texttt{X}, \texttt{Y}) are
  mandatory. $\set{P}$ has to be compliant with the concept hierarchy
  shown in Fig.~\ref{fig:ontology}. $\set{P}_s \subseteq \set{P}$,
  with $|\set{P}_s| > 0$, contains the predicates that provide an
  abstraction of the elements in $\set{M}_s$.
\end{itemize}

Note that the definition of a unique reference frame $R$ allows to
associate the elements of the subset $\set{M}_s$ with those of
$\set{P}_s$. Moreover, the requirement that $\set{M}$ is composed of
geometrical elements obtained as raw sensor data, gives the
opportunity to define an additional functionality on top of our
representation. Indeed, as we will explain in
Section~\ref{sec:evaluation}, we are interested in the possibility to
get the actual sensor data, given a specific pose in the map expressed
according to $R$. For what concerns $\set{P}$, instead, the predicates
\textit{is-a} and \textit{instance-of} represent respectively: the
subclass relation, meaning that if \textit{is-a}(\texttt{B},
\texttt{A}) holds, the class \texttt{B} is a subclass of the class
\texttt{A} and every instance of \texttt{B} is also an instance of
\texttt{A}; the membership relation, meaning that if
\textit{instance-of}(\textit{\texttt{a}}, \texttt{A}) holds, the
individual \texttt{\textit{a}} belongs to the class
\texttt{A}. Additionally, some predicates can have a function-like
behavior, meaning that they can occur only once for each
individual. For example, if dealing with the classes \texttt{Person}
and \texttt{IDNumber}, the predicate \textit{hasId}(\texttt{X},
\texttt{Y}) occurs only once for each instance of \texttt{Person} and
\texttt{IDNumber}.

To give a general idea, let us suppose we are building a semantic map
for a robot operating and interacting with people in a mall. In this
case, we can use our representation and choose $\set{M}$ to be a set
of points, like a unique point cloud modeling the 3D map of the
environment. For what concerns $\set{P}$, we can extend the concept
hierarchy of Fig.~\ref{fig:ontology} as follows:

\begin{itemize}
\item being a person an element of interest, we can define a class
  \texttt{Person} and add the predicate \textit{is-a}(\texttt{Person},
  \texttt{Physical\_Thing});
\item a specialization of the class \texttt{Location} can be
  introduced for the shops and corridors, by defining the classes
  \texttt{Shop}, \texttt{Corridor} and adding the predicates
  \textit{is-a}(\texttt{Shop}, \texttt{Location}),
  \textit{is-a}(\texttt{Corridor}, \texttt{Location});
\item a \texttt{Connecting\_Architecture} can be specified in such a
  way that it always \textit{connects} an element of the class
  \texttt{Shop} and one of the class \texttt{Corridor};
\item since a shop could use advertisements for promoting itself, we
  can define a class \texttt{Advertisement}, add the predicate
  \textit{is-a}(\texttt{Advertisement}, \texttt{Abstract\_Thing}) and
  define a new predicate \textit{hasAdvertisement}(\texttt{X},
  \texttt{Y}), where \texttt{X} could be an instance of \texttt{Shop}
  and \texttt{Y} an instance of \texttt{Advertisement}.
\end{itemize}

Finally, we can select as reference frame $R$ the global frame of a 3D
map.

%%%%%%%%%%%%%%%%%%%%%%%%%%%%%%%%%%%%%%%%%%%%%%%%%%%%%%%%%%%%%%%%%%%%%%%%%%%%%%%%

\section{Semantic Map Evaluation}
\label{sec:evaluation}

Once we are given the representation schema presented in
Section~\ref{sec:representation}, a metric and one \emph{shared}
environment, then it is possible to perform a comparison between two
different methods on the basis of the semantic maps they generate. For
this reason, we have to define one or more metrics that allow for a
quantitative evaluation of each method. Then, we have to find an
environment in which to perform this kind of experiments. While some
Robotics Innovation Facilities
exist\footnote[3]{\url{http://www.echord.eu/facilities-rifs/}} to this
purpose, it is still not easy to retrieve common locations and
environments, mainly due to logistic, physical and economic
constraints. For these reasons, while hypothesizing some metric in
Section~\ref{subsec:metric}, we suggest the construction of a dataset
of semantic maps according to the proposed representation schema. In
particular, the set of geometrical elements $\set{M}$ should be built
with real sensor data. In this way, it is possible to simulate the
robot navigation, as well as its sensor acquisition. This can be done
by defining a projection function that transforms the elements of
$\set{M}$ into the associated sensor domain. For example, in the case
of a RGB-D camera the geometrical elements are projected in a depth
and RGB image, while in the case of a laser they are projected into a
vector of range values.

Such a dataset is a ground truth of each environment and therefore it
can be used to make comparisons based on specific metrics. Of course,
the set $\set{P}$ cannot be fully satisfactory, since it is not
feasible to take into account all the possible semantic knowledge. For
this reason, it is likely that a user might need to extend it. In this
case, it is important to update the original ground truth so that it
becomes more and more complete and that everyone can test their system
on the same dataset.

%%%%%%%%%%%%%%%%%%%%%%%%%%%%%%%%%%%%%%%%%%%%%%%%%%%%%%%%%%%%%%%%%%%%%%%%%%%

\subsection{Evaluation Metric Hypotheses}
\label{subsec:metric}

In this section, we hypothesize some possible evaluation metrics to be
used for comparison between two semantic maps which are compliant with
our previous proposal. Given a representation $\set{SM}_1 = \langle
R_{GT},\set{M}_1,\set{P}_1 \rangle$ and the ground truth
$\set{SM}_{GT} = \langle R_{GT},\set{M}_{GT},\set{P}_{GT} \rangle$, an
evaluation metric can be defined as

\begin{equation}
  \delta(\set{SM}_1,\set{SM}_{GT}) = f(|\set{M}_1 \ominus \set{M}_{GT}|, |\set{P}_1 \boxminus \set{P}_{GT}|).
\end{equation}

Note that the reference frame $R_{GT}$ of $\set{SM}_1$ and
$\set{SM}_{GT}$ coincide: this is easily achievable by applying the
transformation offset between the original frame $R_1$ of $\set{SM}_1$
and $R_{GT}$ of $\set{SM}_{GT}$. The definition of the operators
$\ominus$ and $\boxminus$ determines the metric itself. For example,
$\ominus$ can be a distance $d$ between geometrical elements,
according to Table~\ref{tab:spatial_metric}, while the $\boxminus$
operator could return two sets of predicates $\Delta$ and $\Gamma$
such that:

\begin{equation}
  \label{eq:def-metric}
  \{\set{P}_1 \setminus \Gamma\} \cup \Delta \models \set{P}_{GT}
\end{equation}

\begin{table}[b!]
  \centering
  \caption{\scriptsize Example definition of the $\ominus$ operator. The index 
    $i$ indicates the $i$-th corresponding geometric element in 
    $\set{M}_1$ and $\set{M}_2$, while $p$, $l$ and $\pi$ represent 
    respectively a point, a line and a plane.}
  \label{tab:spatial_metric}
  \setlength{\tabcolsep}{.5em}
  \begin{tabular}{ccccc}
    & & \multicolumn{3}{c}{$\set{M}_{GT}$} \\ \hhline{~~===} 
    & & Points & Lines & Planes \\ \cline{3-5} 
    & & & & \\ [-6pt]
    \multicolumn{1}{c|}{\multirow{3}{*}{$\set{M}_1$}} & \multicolumn{1}{||c|}{Points} & \makecell{$\sum_{i}{d(p_{i}^1,p_{i}^{GT})}$}   & \makecell{$\sum_{i}{d(p_{i}^1,l_{i}^{GT})}$}   & \makecell{$\sum_{i}{d(p_{i}^1,\pi_{i}^{GT})}$}   \\ [5pt]
    \multicolumn{1}{c|}{}                             & \multicolumn{1}{||c|}{Lines}  & \makecell{$\sum_{i}{d(l_{i}^1,p_{i}^{GT})}$}   & \makecell{$\sum_{i}{d(l_{i}^1,l_{i}^{GT})}$}   & \makecell{$\sum_{i}{d(l_{i}^1,\pi_{i}^{GT})}$}   \\ [5pt]
    \multicolumn{1}{c|}{}                             & \multicolumn{1}{||c|}{Planes} & \makecell{$\sum_{i}{d(\pi_{i}^1,p_{i}^{GT})}$} & \makecell{$\sum_{i}{d(\pi_{i}^1,l_{i}^{GT})}$} & \makecell{$\sum_{i}{d(\pi_{i}^1,\pi_{i}^{GT})}$} \\ 
  \end{tabular}
\end{table}

The lower the cardinality of $\Delta$ and $\Gamma$, the better is the
semantic representation. However, this does not consider the fact that
the subset $\set{P}_s$ contains some reference to spatial information
(which could be measured again by metric criteria). A solution to this
problem could be the redefinition of $\boxminus$ as an operator which
returns two sets of predicates $\Delta$ and $\Gamma$, and a distance
$d$ such that:

\begin{equation}
  \{(\set{P}_1 \setminus \set{P}_{1_s}) \setminus \Gamma\} \cup \Delta \models \{\set{P}_{GT} \setminus \set{P}_{GT_s}\},~~
  d(\set{P}_{1_s}, \set{P}_{GT_s}).
\end{equation}

For example, suppose that the ground truth $\set{SM}_{GT}$ contains a
table and a chair correctly positioned. If the table is missing in the
set $\set{P}_1$ of the robot semantic map $\set{SM}_{1}$, from our
metric in Eq.~\ref{eq:def-metric} we obtain that $\Delta$ has
cardinality $1$. Indeed, in this case the robot would not be able to
execute the command ``go to the table''. Conversely, if the table
belongs to $\set{P}_s$, the cardinality is $0$ and the robot is able
to execute the command. Similarly, if the object is not well
positioned in $\set{M}_1$ any distance from
Table~\ref{tab:spatial_metric} would be much bigger than zero, and the
robot would execute the command by reaching a wrong
location. Additional metrics could be defined on different criteria
like the processing time, the distance traveled by the robot, the
number of sensor readings processed, etc.

%%%%%%%%%%%%%%%%%%%%%%%%%%%%%%%%%%%%%%%%%%%%%%%%%%%%%%%%%%%%%%%%%%%%%%%%%%%%%%%%

\section{Dataset Construction}
\label{sec:dataset}

\begin{figure}[t!]
  \centering
  \includegraphics[width=0.9\columnwidth]{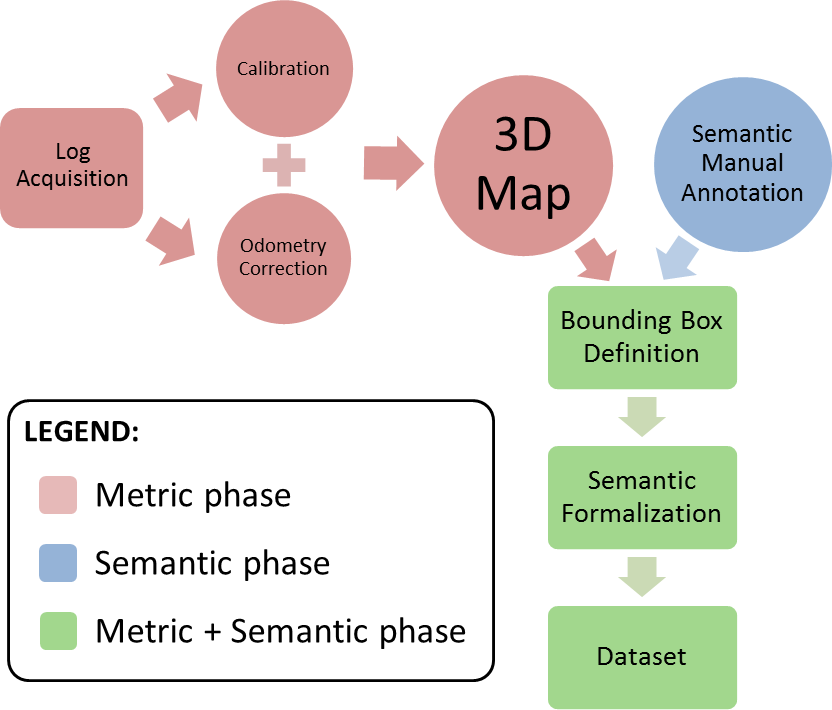}
  \caption{Steps involved in the process of building the dataset.}
  \label{fig:methodology}
\end{figure}

Since the construction of the dataset is based on the representation
proposed in Section~\ref{sec:representation}, and it consists of the
combination of spatial and semantic information, any approach
compliant with that could be applied. In this section we describe our
method for the generation of a ground truth, in which the set
$\set{M}$ consists of a 3D point cloud, $\set{P}$ implements the
proposed concept hierarchy and $\set{P}_s$ contains abstractions of
bounding boxes. In particular, in order to collaborate with a larger
community of researchers, we consider low cost sensors (i.e., RGB-D
cameras like Microsoft Kinect and Asus Xtion) which can be easily
found on any robot. Note that building a 3D map with this kind of
sensors, leads to multiple open issues. Still, even if with an
additional manual refinement, our software allows to build such
maps. As shown in Fig.~\ref{fig:methodology}, this process is composed
of several steps, which can be divided into metric and semantic
phases. First, we acquire data in order to generate a 3D map and we
perform a preliminary manual annotation of the objects inside the
environment. Then, by associating semantic information and volumes in
the 3D map, in the form of bounding boxes, we obtain the desired
semantic map. Of course, sensor calibration prior to data acquisition
is highly recommended (see Section~\ref{subsec:calibration} for more
details).

%%%%%%%%%%%%%%%%%%%%%%%%%%%%%%%%%%%%%%%%%%%%%%%%%%%%%%%%%%%%%%%%%%%%%%%%%%%

\subsection{Data Acquisition}
\label{subsec:acquisition}

The data acquisition step can be divided in two different parts, one
related to the 3D map, the other to the semantic annotations for
elements of interest inside the environment. While manually collecting
semantic annotations is relatively easy, although tedious, 3D data
acquisition results to be more challenging due to the limitations of
low cost sensors. 

The generation of a 3D map requires the acquisition of a log capturing
the income of the robot sensors while moving around the
environment. In particular, this should contain the robot odometry (or
laser data) and the camera stream (both for depth and RGB). While
taking the log, one should pay attention to steer the robot so that at
least one camera does not see only a flat surface. Indeed, structures
like a floor, a wall or two parallel planes do not help the mapping
system, due to their poor geometrical information.

%%%%%%%%%%%%%%%%%%%%%%%%%%%%%%%%%%%%%%%%%%%%%%%%%%%%%%%%%%%%%%%%%%%%%%%%%%%

\subsection{Sensor Calibration}
\label{subsec:calibration}

The calibration of a sensor is the process of correctly
computing its internal parameters, as well as its pose with respect to
the robot reference frame. Extracting the right internal parameters
improves the data generated by the sensor reducing its intrinsic
error. For example, in the case of a depth camera, this corresponds to
determine its camera matrix and distortion parameters. Computing the
correct pose of a sensor, instead, allows to accurately express data
measurements with respect to a different reference frame.

In order to perform sensor calibration and supposing to use $n$ RGB-D
cameras on the robot, $n + 2$ logs\footnote{A log is obtained by
  acquiring and recording the required sensor data.} are required. In
particular, choosing one of the cameras as a reference, we have:

\begin{enumerate}
\item \label{log:internal} $n$ \emph{intrinsic calibration logs},
  containing the stream of the $i$-th RGB-D sensor, for the
  calibration of the internal parameters of its depth camera (refer
  to~\cite{Dicicco2014} for more details on how to acquire data);
\item \label{log:sensor_base} $1$ \emph{sensor-base calibration log},
  containing the robot odometry (or laser data) and the camera stream,
  for calculating the pose between the robot and reference RGB-D
  sensor (the robot should slowly translate and rotate while the
  reference sensor sees at least 3 planes, each of them being non
  parallel with all the others);
\item \label{log:sensor_sensor} $1$ \emph{sensor-sensor calibration
    log} (at least), containing the stream of the $n$ cameras, for
  computing the pose of $n-1$ RGB-D sensors with respect to the
  reference one (all the cameras should see, at least once, the same
  part of the environment while \emph{always} respecting the condition
  of the previous point);
\end{enumerate}

Common RGB-D cameras are affected by a substantial distortion in the
depth channel. Not considering this distortion leads to systematic
drifts in the estimate of the robot pose while mapping.  This
calibration is performed by following the procedure explained by Di
Cicco \emph{et al.}~\cite{Dicicco2014} on the intrinsic calibration
logs. At the end of this procedure, it is possible to reduce the
intrinsic error which normally affects the sensors data (i.e., walls
that should be flat, look curved on the edges).

Another goal of the calibration procedure is to find the pose of one
of the cameras (\emph{reference}) with respect to the robot frame, and
the relative offsets (translation and rotation) between all the other
cameras and the \emph{reference}. The software we developed provides
two different tools to compute these offsets. The first one performs
the computation of the transform $\mathbf{T^*}$ between the robot
frame and the reference depth camera. By using the sensor-base
calibration log we estimate the motion of the camera in a small
region. Taking as reference the odometry of the robot, this tool casts
a least square problem that minimizes a cost function which depends on
the sensor transform $\mathbf{T}$ and returns $\mathbf{T^*}$. The
second tool, instead, allows the computation of the offset between
pairs of depth cameras. The main idea is to use the sensor-sensor
calibration log to generate, for each camera, an independent point
cloud. In this way, each sensor produces a cloud starting from its own
reference frame. Once this is done, our registration algorithm can be
run between pairs of point clouds. The output of the alignment
determines the relative translation and rotation between the origins
of the point clouds and thus between the sensors.

\begin{figure}[t!]
\centering
\includegraphics[width=0.7\columnwidth]{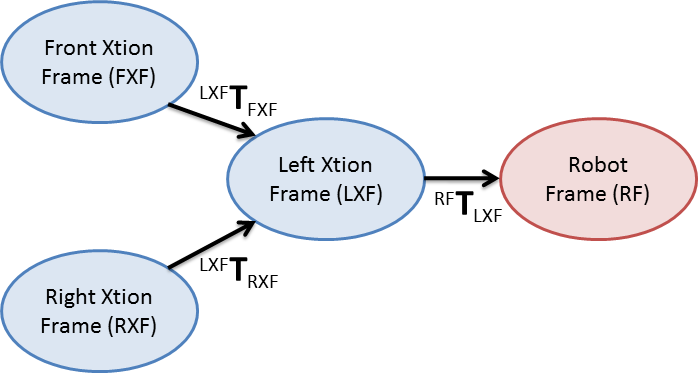}
\caption{Sensor transformation tree generated at the end of a
  calibration procedure. In this case the robot was equipped with 3
  depth cameras.}
\label{fig:transformations_tree}
\end{figure}

At the end of the calibration we are able to construct a tree of
sensor pose transformations (see
Fig.~\ref{fig:transformations_tree}). From this tree, it is possible
to compute the transformation between any two nodes, by a simple
offset concatenation.

%%%%%%%%%%%%%%%%%%%%%%%%%%%%%%%%%%%%%%%%%%%%%%%%%%%%%%%%%%%%%%%%%%%%%%%%%%%

\subsection{Data Processing}
\label{subsec:processing}

% \begin{figure}[t!]
% \centering
% \includegraphics[width=\columnwidth]{images/local_map.png}
% \caption{Example of a local map and its internal trajectory. }
% \label{fig:local_map}
% \end{figure}

Once all the data is acquired, the 3D map can be built. To this end,
the point clouds recorded in the log are aligned generating a set of
\emph{local maps}. A local map is a point cloud constructed by
aligning and integrating a sequence of depth sensor data while the
robot moves in the environment. This is obtained through the use of a
point cloud registration algorithm based on the work by Serafin
\emph{et al.}~\cite{Serafin2014}. A new local map is started whenever
one of the two following statements holds:

\begin{itemize}
\item the estimate of the robot (or equivalently the camera) movement
  is greater than a certain amount. This allows to limit the growth of
  the local map in terms of dimension;
\item the point cloud registration algorithm detects that the last
  alignment is not good (with possibility of inconsistency). This is
  necessary in order to avoid to introduce errors inside the local map.
\end{itemize}

The local map generator uses the robot odometry as initial guess for
the point cloud alignment. However, a good odometry estimation is not
always available. In this case (but this is useful in general), if the
robot comes with a 2D laser, it is possible to use as initial guess
the transformation provided by the \emph{scan matcher} developed as
part of our software. The 3D map is represented as a pose
graph~\cite{Grisetti2010}, where each local map is connected to the
previous and following one by means of a transformation. More in
detail, nodes of the pose graph represent local maps, with their
position and orientation in a global frame. Edges, instead, are
relative transforms between local maps. The benefits of this metric
representation are that it allows to add/remove anytime information
and update an existing map. Indeed, by using a tool provided in our
software, inconsistencies in the map can be manually fixed. More
specifically, the user can select and align two nodes of the graph at
time and add a new edge between them. This, together with the
optimization of the pose graph \cite{Kummerle2011}, leads to the
elimination of inconsistencies and thus, to a refined map.

%%%%%%%%%%%%%%%%%%%%%%%%%%%%%%%%%%%%%%%%%%%%%%%%%%%%%%%%%%%%%%%%%%%%%%%%%%%

\subsection{Combining 3D Map and Semantic Data}
\label{subsec:fusion}

Once both the 3D map and the semantic annotations are available it is
possible to combine them by means of a geometric abstraction like a
volume in the map. In our case, we define such a volume to be a
bounding box (i.e., a parallelepiped) containing all the geometric
elements to which we want to attach the same semantic information.

After all the bounding boxes are assigned, we formalize the predicates
$\set{P}$ (compliant with the conceptual hierarchy) in OWL-DL, by
using
Prot\'{e}g\'{e}~\footnote[4]{\url{http://protege.stanford.edu/}}. Bounding
boxes, in particular, belong to the subset $\set{P}_s$ and they are
formalized by means of classes like \texttt{Size}, \texttt{Position}
and \texttt{Shape}.

%%%%%%%%%%%%%%%%%%%%%%%%%%%%%%%%%%%%%%%%%%%%%%%%%%%%%%%%%%%%%%%%%%%%%%%%%%%

\subsection{Dataset Example}
\label{subsec:example}

\begin{figure}[t!]
  \centering
  \includegraphics[width=0.85\columnwidth]{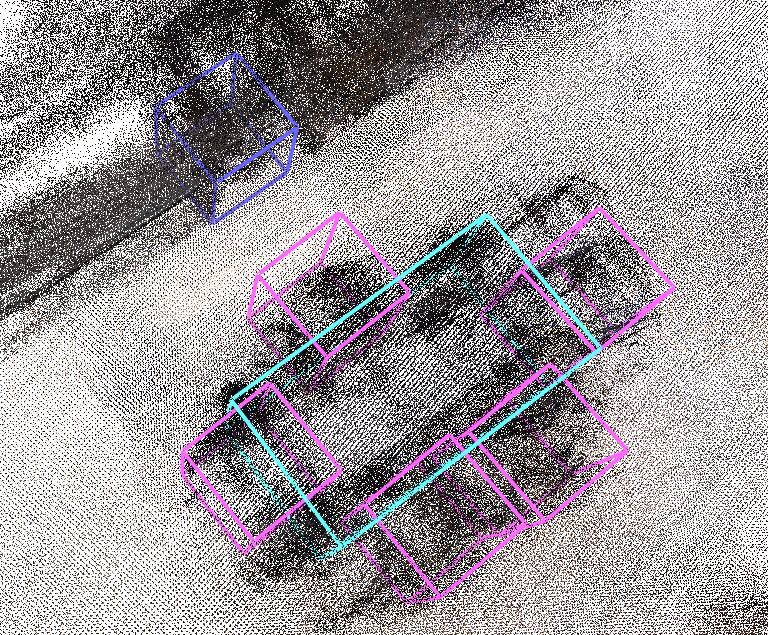}
  \caption{Detail of the example dataset acquired in the RIF of
    Peccioli. The image shows a table and chairs with their associated
    bounding boxes. RGB information is intentionally omitted and
    resolution is reduced for a better visualization of the bounding
    boxes.}
  \label{fig:example}
\end{figure}

We performed the procedure described so far on a set of data
specifically acquired during the RoCKIn
Camp\footnote[5]{\url{http://rockinrobotchallenge.eu/}} held in the
ECHORD++ Robotic Innovation Facility of
Peccioli\footnote[6]{\url{http://www.echord.eu/facilities-rifs/the-peccioli-rif/}},
in Italy. In particular, this is a domestic environment with several
rooms and everyday objects built to foster benchmarking of robotic
applications, to test their robustness, and to support standardization
efforts. While a detail of the 3D map of the environment is shown in
Fig.~\ref{fig:example}, the whole dataset is hosted online
(\url{http://goo.gl/v7xSyl}) and contains a ground truth
representation which is compliant with the requirements stated in
Section~\ref{sec:representation}. Namely, a 3D point cloud with an
associated reference frame and the corresponding OWL-DL ontology
compose the first example of a dataset for semantic maps.

%%%%%%%%%%%%%%%%%%%%%%%%%%%%%%%%%%%%%%%%%%%%%%%%%%%%%%%%%%%%%%%%%%%%%%%%%%%%%%%%

\section{Discussion}
\label{sec:discussion}
In this paper we defined a methodology for representing semantic
maps. In particular, we designed a formalization of their
representation which includes both spatial and semantic knowledge. On
top of this, we made some hypotheses for metrics and evaluation
criteria, based on the idea that a ground truth for semantic maps
exists. Note that the procedure we proposed for building a dataset is
based on real sensor data. This allows to simulate robot navigation
inside the environment, breaking down logistic, physical and economic
barriers for a fair comparison between different semantic mapping
methods. Finally, we provided useful documented open-source software
for building such a dataset (\url{http://goo.gl/v7xSyl}). We invite,
in this way, the scientific community to contribute in populating the
dataset with more and more annotations and environments. In addition
to all of this, we have also shown a first real example of ground
truth for a semantic map.  Open challenges, however, still
remains. Future work, for example, should be oriented to the
definition of a standard metric of evaluation.

%%%%%%%%%%%%%%%%%%%%%%%%%%%%%%%%%%%%%%%%%%%%%%%%%%%%%%%%%%%%%%%%%%%%%%%%%%%%%%%%

\bibliographystyle{IEEEtran}
\bibliography{references}

\balance
\end{document}